\documentclass{article}

\usepackage[final,nonatbib]{nips_2018}
 \usepackage[pdftex]{graphicx}

\usepackage[utf8]{inputenc} 
\usepackage[T1]{fontenc}    
\usepackage{hyperref}       
\usepackage{url}            
\usepackage{booktabs}       
\usepackage{amsfonts}       
\usepackage{nicefrac}       
\usepackage{microtype}      

\title{Addressing the Invisible: Street Address Generation for Developing Countries with Deep Learning}

\author{
 Ilke Demir \\
  Facebook\\
  \And
  Ramesh Raskar \\
  MIT Media Lab \\
}

\begin{document}

\maketitle

\begin{abstract}
More than half of the world's roads lack adequate street addressing systems. Lack of addresses is even more visible in daily lives of people in developing countries. We would like to object to the assumption that having an address is a luxury, by proposing a generative address design that maps the world in accordance with streets. The addressing scheme is designed considering several traditional street addressing methodologies employed in the urban development scenarios around the world. Our algorithm applies deep learning to extract roads from satellite images, converts the road pixel confidences into a road network, partitions the road network to find neighborhoods, and labels the regions, roads, and address units using graph- and proximity-based algorithms.
We present our results on a sample US city, and several developing cities, compare travel times of users using current ad hoc and new complete addresses, and contrast our addressing solution to current industrial and open geocoding alternatives.
\end{abstract}

\section{Introduction}
Street addresses enhance precise physical presence and effectively increase the connectivity all around the world. Currently 75\% of the roads in the world are not mapped~\cite{w3w}, and this number is increasing in developing countries. United Nations claims to have 4 billion invisible people in the world due to the addressing problem in developing countries. This problem is even more critical in disaster zones, areas with limited resources, and geographically challenging locations. As the remote sensing technology has been significantly improving over the past decade, the organic growth of urban and suburban areas outruns the deployment of addressing schemes. We want to address the invisible. We imagine an algorithm that automatically creates meaningful addresses for unmapped areas, areas with no street name or address. 



In order to realize our goal, we designed and implemented a generative addressing system to bridge the gap between grid-based digital addressing schemes and traditional street addresses. We (i) design a physical addressing scheme, which is linear, hierarchical, flexible, intuitive, perceptible, robust, (ii) propose a segmentation method to obtain road segments and regions from satellite imagery, using deep learning and graph-partitioning, (iii) implement a labeling method to name urban elements based on current addressing schemes and distance fields, and (iv) develop ready-to-deploy prototype applications supporting forward and inverse geoqueries.

\section{Related Work}

\textbf{Geocoding approaches:} Popular geocoding solutions are either not in human-readable form (e.g., GooglePlaceID and OkHi), or tend to de-correlate from the topological structure (e.g., \cite{w3w}, Zippr and MapTags), and they lack essential properties of a street addressing system, such as linearity and hierarchy. Those properties as well as their human intersection become even more crucial in developing countries ~\cite{mitblog}. 

\textbf{Procedural generation:} Automating the generation of maps is extensively studied in the urban procedural modeling world ~\cite{chen2008,vanegas2012,parish2001,aliaga2008,sun2002}, creating detailed and structurally realistic models, but none of them are applicable as a synthesis method on real world. On the other hand, inverse procedural modeling approaches~\cite{aliaga2016} process real-world data for generative representations. We follow this last path and rely on satellite imagery as the input of our synthesis approach. 

\textbf{Remote sensing:} Several approaches automate the extraction of geospatial information using already existing data resources ~\cite{wang2013,skoumas2016,mattyus2015,deeproadmapper,zeng2017}. Similar approaches extract road networks using neural networks~\cite{wang2015,zhao2012,li2016,xu2014,peteri2003, poullis2008,demir18}.Processing the road topology has also been studied by clustering and graph partitioning approaches~\cite{wegner2013,alshehhi2017,anwar2017}. 


\textbf{Addresses around the world:} Traditional street addresses and names are usually the result of cultural dynamics, politics, economies, and other long-term processes adopted by urban authorities. We conducted some research on the current addressing methods such as the London postal code system~\cite{London}, South Korea street naming, Berlin house numbering, and more~\cite{streetbook}. 

\section{The Street Address Format}\label{sec3}

\begin{figure}[h]
\begin{center}
   \includegraphics[width=0.8\linewidth]{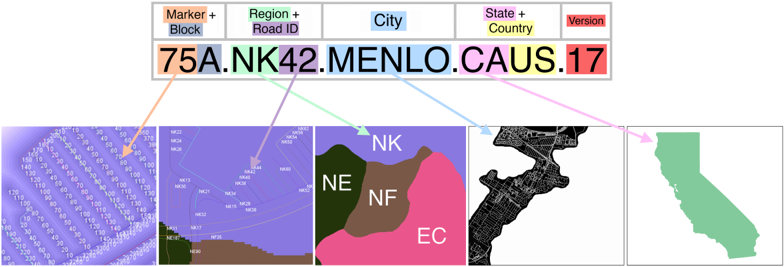}
\end{center}\vspace{-12pt}
   \caption{Street Address Format: house number, road name, city, state (if~applicable), country, followed by the version year.}
\label{fig:add2}
\end{figure}

Semantic properties emphasize user-friendly features, implying linear, hierarchical, universal,
and memorable addresses. Structural properties enable the format of street addresses to be computer friendly, necessitating linear, hierarchical, compressible, robust, extendible, and queryable codes.  Following these design principles (Figure~\ref{fig:add2}), the last field indicates version, the fourth field contains the country and state information when applicable, preceded by the city information in the third field. The second field contains the road name, which starts with the region label, followed by the road number. Lastly, the first field is composed of the meter marker along the road and the block letter from the road, animating the house number and apartment number consecutively.


\vspace{-6pt}

\section{Our Generative Addressing System}
The segmentation step extracts roads, breaks them into road segments and clusters them into regions. The labeling step names the regions, road segments, and place markers and assigns block letters to individual addressable units. Details of each module can be found in our extended journal paper~\cite{ijgi}.

\begin{figure}[h]
\begin{center}
  \includegraphics[width=0.8\linewidth]{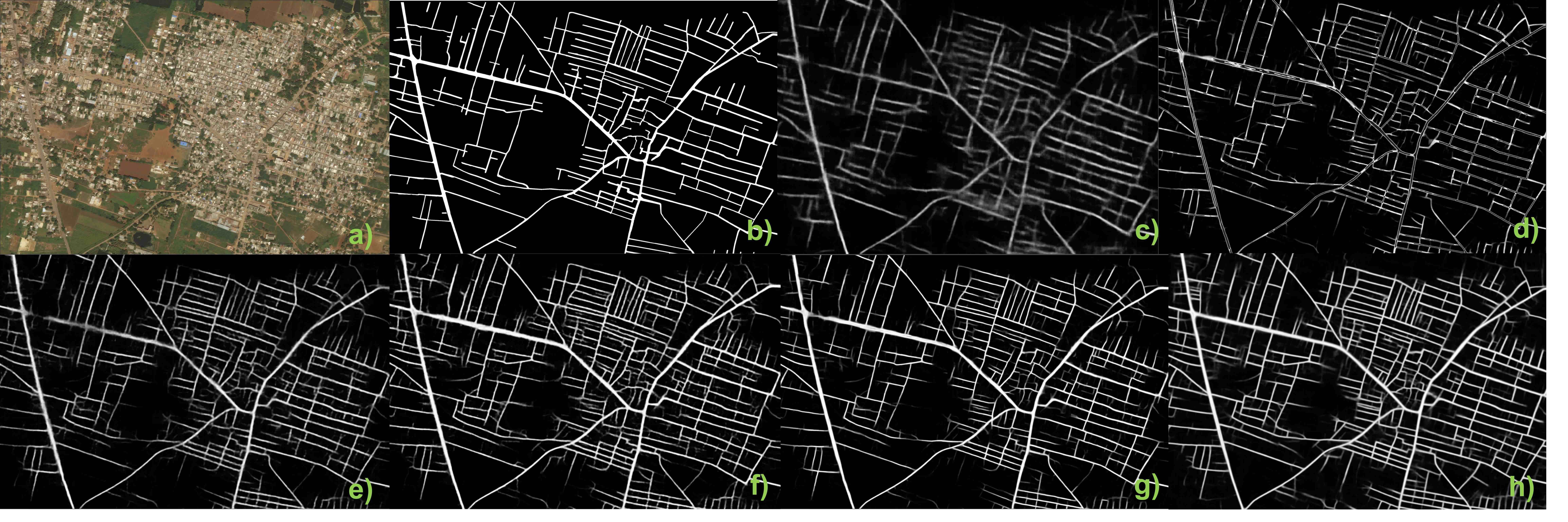}
\end{center}\vspace{-6pt}

   \caption{Comparison of NN Models. An example (\textbf{a}) satellite image and (\textbf{b}) ground truth; and road predictions using (\textbf{c}) VGG; (\textbf{d}) U-Net; (\textbf{e}) ResNet50; (\textbf{f}) ResNet101; (\textbf{g}) SegNet; and (\textbf{h}) DeepLab.}
\label{fig:pred} 
\end{figure}

\textbf{Predicting Road Pixels:}
The first step of our approach creates binary road prediction images from three channel satellite images of 0.5 m resolution and of size \mbox{19 K * 19 K}. Both training and testing are done with patches of $192 \times 192$. The training set includes 4--16 tiles per country, and the test set includes all the rest of the tiles, manually spatially distributed to sample all areas, keeping the ratios mostly at 70\% to 30\%. We use a modified version of SegNet~\cite{segnet}, which consists of the first 13 convolutional layers of the VGG16 network for the encoder, having a corresponding decoder layer for each encoder. We modify the last soft-max layer to change the multi-class structure to have binary classes for road detection, by substituting it with a convolutional layer. We experimented with architectures (Figure~\ref{fig:pred}) such as VGG~\cite{vgg}, U-Net~\cite{unet}, and ResNet~\cite{resnet} variations; however, we achieved the best result with the SegNet model trained on dense and diverse tiles, resulting in 72.6\% precision and 57.2\% recall. We also experimented with DeepLab~\cite{deeplab} variations and achieved 75.4\% precision and 75.9\% recall; however the model showed signs of overfitting after epoch 30. 

\textbf{Creating the road graph:}
The post-processing includes binarizing the image with thresholding, running a depth-first search to join connected roads using the confidences, applying an~orientation-based adaptive median filtering on the road end points, bucketing the road segments based on their orientations, and processing intersections. Then we convert the road segments into a road graph, nodes as intersections, edges as road segments, and weights as the segment distance.

\textbf{Defining regions:}
 We~partition the road graph into communities that have the maximum interconnectivity and minimum intraconnectivity. We experimented with normalized min-cut~\cite{mincut}, Newman--Girvan~\cite{newman}, and~optimal modularity-based~\cite{modularity} graph partitioning approaches, concluding on \cite{mincut}, being accurate and significantly more efficient choice. 

\textbf{Labeling regions, roads, and address units:}
We compute the most dense region by averaging number of roads per unit area, and we name this region \textit{``CA''} for the city center. We divide all other regions into four orientations: \textit{N(orth), S(outh), W(est),} and \textit{E(ast)} and assign letters in that specific order, following the spiral pattern of the London post code system. The roads in each region are divided into two main directions: odd for north--south bound, and even for east--west bound, then numbered according to their order. For each road segment, we place a virtual meter marker every 5 m, on the left and right sides of the roads by even and odd numbering. Finally, we compute a distance field of the roads and discretize that field by a 5 m step size, as the block letter. 


\section{Results and Applications}

Our system is written in Python and C++; the implementation is on the CPU (except road prediction). We use Chainer \cite{chainer}, networkx, and sci-kit libraries. We used our approach to process more than 10 cities, totaling up to more than 16 K\ km$^2$. The source code to convert .osm files and geotiffs to street addresses is available on our repository\footnote{https://github.com/facebookresearch/street-addresses}.

We compare the road predictions with ground truth for the extracted roads of an unmapped suburban area. Our SegNet model and post-processing approach were able to learn 90.51\% of the roads. This success ratio was close to 80\% on average per city.
\begin{figure}[h]
\begin{center}
   \includegraphics[width=0.7\linewidth]{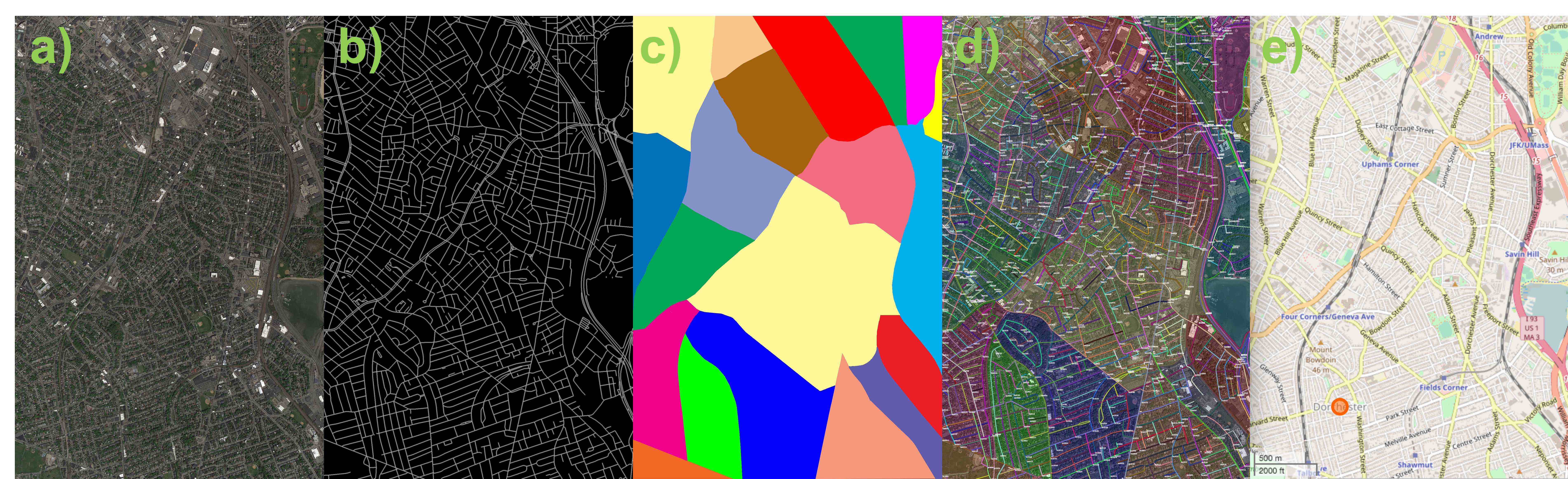}
\end{center}\vspace{-12pt}
   \caption{We show (\textbf{a}) input satellite tile; (\textbf{b}) extracted roads; (\textbf{c}) created regions; and (\textbf{d}) generated map; compared to (\textbf{e}) OpenStreetMap (OSM) of the same area.}
\label{fig:dor}
\end{figure}

We evaluate the usefulness of our generative maps with some treasure hunt-like user experiences. We compared the travel times using the old and new addressing schemes. Overall, the travel times decreased by 21.7\%, with our system producing a 52.4 s improvement on average and decreasing the last mile of activity, proving the accuracy of our addresses.

We compare street segments dictated by the traditional addresses and our generative addresses (Figure~\ref{fig:dor}). Comparing the road segments, we accomplished extracting 95\% of the roads in that particular city tile. Comparing the addresses, although traditional addresses are more established, our addresses are easier to remember and support intuitive self-location and navigation.
\begin{figure}[h]
\begin{center}
   \includegraphics[width=0.9\linewidth]{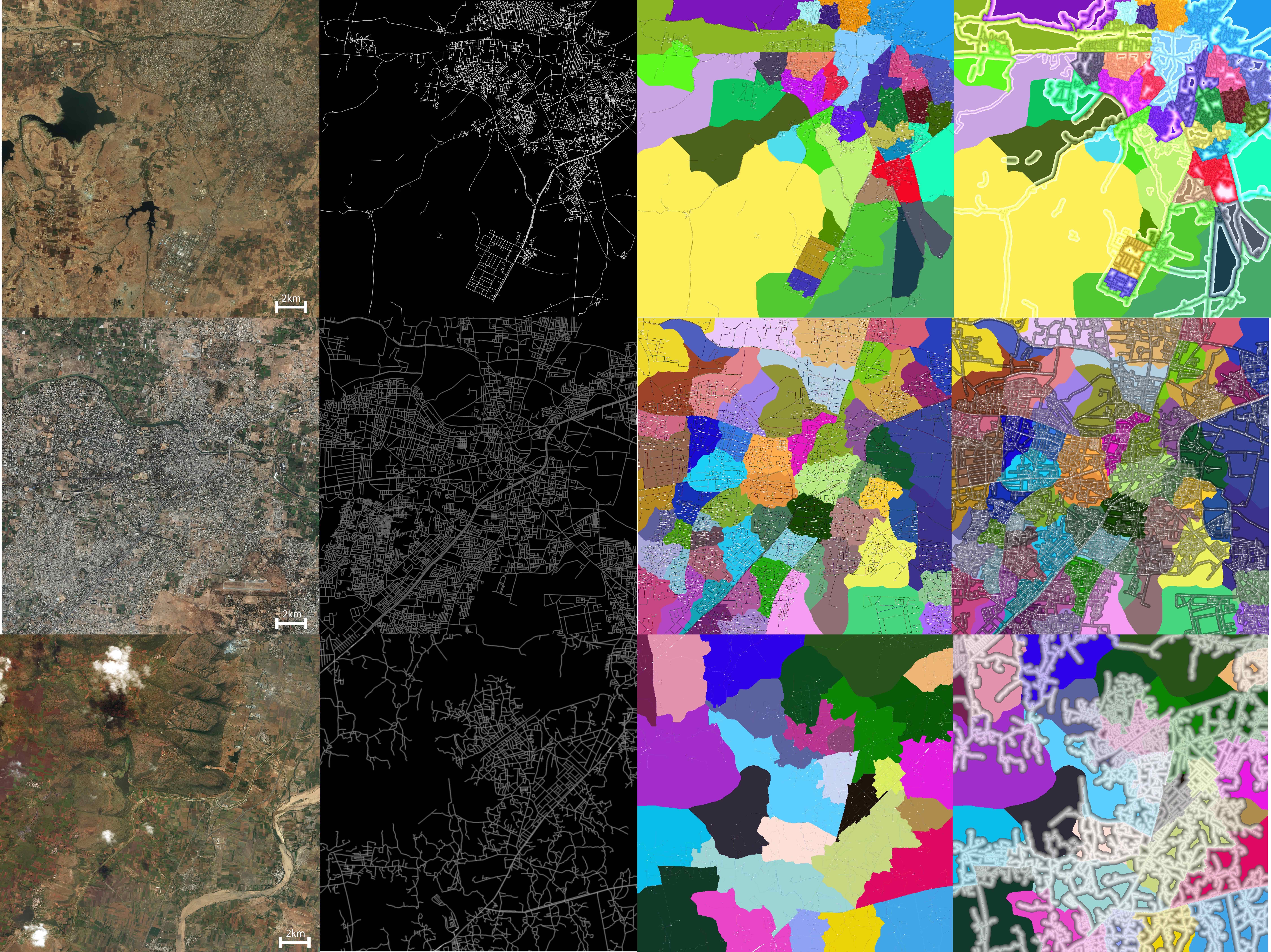}
\end{center}\vspace{-12pt}
   \caption{Satellite image, extracted roads, labeled regions and roads, and meter markers and blocks of three example developing cities.}
\label{fig:india}
\end{figure}

However, keeping the motivation of providing street addresses to the approximately 4 billion unconnected people, our results in fact shine for developing countries. Figure~\ref{fig:india} shows our generative maps in the same format, on three different unmapped developing cities. We accomplished automatically addressing more than 80\% of the populated areas, which significantly improved map coverage.

We compare our maps to other popular addressing solutions. For the same point on earth, \cite{w3w} outputs three random words \textit{parrot.failed.casino}, Google Maps contains some unlabelled roads; however it outputs \textit{Green Park} for a couple of kilometers around the point. OSM does not  contain roads, and the point can be reached only by its latitude and longitude. However, our approach extract the roads almost completely and assign a unique address as \textit{715D.NE127.Dhule.MhIn}.

\section{Conclusions}
We demonstrated that deep learning can be used in a world-wide system for automatically creating street addresses from satellite imagery for developing countries in the world. Physically connecting the unconnected populations should increase the economic, juridical, and life-sustaining involvement of people all around the world. It improves the outreach of businesses and the economy, as well as the accuracy and efficiency of providing first aid in disaster zones. More evaluation, results, and details about determinism and complexity analysis of submodules, constructing a global address space, versioning and updating, handling missing city boundaries, overflowing regions, and 3D roads can be found in our previously published works~\cite{ijgi,cvpr17}.

{\small
\bibliographystyle{ieee}
\bibliography{egbib}
}

\end{document}